\documentclass[letterpaper]{article} 
\usepackage{aaai2026}  
\usepackage{times}  
\usepackage{helvet}  
\usepackage{courier}  
\usepackage[hyphens]{url}  
\usepackage{graphicx} 
\usepackage{natbib}  
\usepackage{caption} 
\usepackage{algorithm}
\usepackage{algorithmic}

\usepackage{newfloat}
\usepackage{listings}
\DeclareCaptionStyle{ruled}{labelfont=normalfont,labelsep=colon,strut=off} 
\floatstyle{ruled}
\newfloat{listing}{tb}{lst}{}
\floatname{listing}{Listing}

\title{FT-NCFM: An Influence-Aware Data Distillation Framework for Efficient VLA Models}
\author{
	\textbf{Kewei Chen}\textsuperscript{\rm 1, 2},
	\textbf{Yayu Long}\textsuperscript{\rm 1, 2},
	\textbf{Shuai Li}\textsuperscript{\rm 3},
	\textbf{Mingsheng Shang}\textsuperscript{\rm 1, 2}\thanks{Corresponding author.}
}
\affiliations{
	\textsuperscript{\rm 1}Chongqing Institute of Green and Intelligent Technology, Chinese Academy of Sciences\\
	\textsuperscript{\rm 2}Chongqing School, University of Chinese Academy of Sciences\\
	\textsuperscript{\rm 3}Faculty of Information Technology and Electrical Engineering, University of Oulu, Finland\\
	\{chenkewei24, longyayu24\}@mails.ucas.ac.cn, shuai.li@oulu.fi, msshang@cigit.ac.cn
}

\usepackage[table]{xcolor}
\usepackage{booktabs}  
\usepackage{multirow}  
\usepackage{amssymb}   
\usepackage{algorithm}
\usepackage{algorithmic}
\usepackage{amsmath}
\usepackage{amssymb}    
\usepackage{algorithm}
\usepackage{algorithmic}
\usepackage{newfloat}
\usepackage{listings}
\DeclareCaptionStyle{ruled}{labelfont=normalfont,labelsep=colon,strut=off} 
\floatstyle{ruled}
\newfloat{listing}{tb}{lst}{}
\floatname{listing}{Listing}

\usepackage{aaai2026}
\usepackage{times}
\usepackage{helvet}
\usepackage{courier}
\usepackage[hyphens]{url}
\usepackage{graphicx}
\usepackage{natbib}
\usepackage{caption}
\title{FT-NCFM: An Influence-Aware Data Distillation Framework for Efficient VLA Models}
\nocopyright
\begin{document}
	
	\maketitle
	
	\begin{figure*}[h!]
		\centering
		\includegraphics[width=\textwidth]{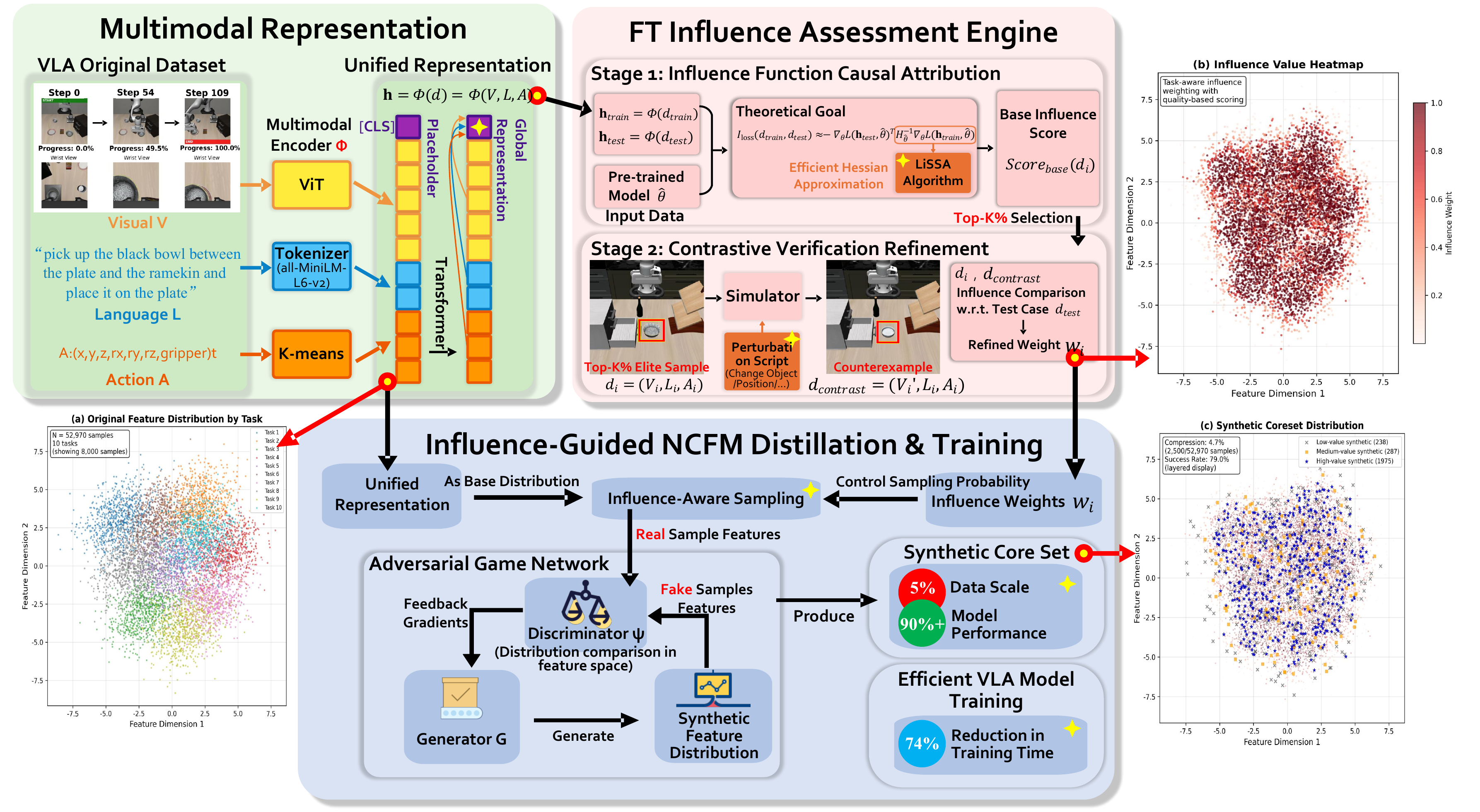} 
		\caption{Overview of our proposed FT-NCFM framework. The framework follows a three-stage pipeline to distill a high-value synthetic coreset from large VLA datasets for efficient robot policy learning.
			\textbf{(Top Left) Multimodal Representation Module}: Converts raw VLA data streams into unified token sequences through respective encoders. A Transformer backbone fuses them into a global feature representation \textbf{h}.
			\textbf{(Top Right) FT Influence Assessment Engine}: A two-stage value assessment process. Stage one uses influence functions for causal attribution to calculate base influence scores. Stage two performs contrastive verification on top-K\% elite samples by programmatically generating "minimal counterexamples" in a simulator, refining their influence weights \textbf{W}.
			\textbf{(Bottom) Influence-Guided NCFM Distillation}: The weights \textbf{W} guide an adversarial network through influence-aware sampling. The discriminator \textbf{$\Psi$} contrasts "weighted real sample features" with "synthetic features" and provides feedback gradients to the generator \textbf{G}. \textbf{G} produces a synthetic coreset for efficient training of downstream VLA models. The t-SNE plots (a), (b), and (c) further visualize this process: (a) shows the original feature distribution by task (different colors represent different tasks); (b) shows the influence value heatmap after FT assessment (color intensity indicates influence weights); (c) clearly demonstrates that our method's synthetic coreset successfully covers the feature distribution of the original high-value samples, reflecting higher information density, with samples categorized into high-value (blue, 1975), medium-value (orange, 287), and low-value (gray, 238) regions.}
		\label{fig:framework_overview} 
	\end{figure*}
	
	\begin{abstract}
		The powerful generalization of Vision-Language-Action (VLA) models is bottlenecked by their heavy reliance on massive, redundant, and unevenly valued datasets, hindering their widespread application. Existing model-centric optimization paths, such as model compression (which often leads to performance degradation) or policy distillation (whose products are model-dependent and lack generality), fail to fundamentally address this data-level challenge. To this end, this paper introduces FT-NCFM, a fundamentally different, data-centric generative data distillation framework. Our framework employs a self-contained Fact-Tracing (FT) engine that combines causal attribution with programmatic contrastive verification to assess the intrinsic value of samples. Guided by these assessments, an adversarial NCFM process synthesizes a model-agnostic, information-dense, and reusable data asset. Experimental results on several mainstream VLA benchmarks show that models trained on just 5\% of our distilled coreset achieve a success rate of 85-90\% compared with training on the full dataset, while reducing training time by over 80\%. Our work demonstrates that intelligent data distillation is a highly promising new path for building efficient, high-performance VLA models.
	\end{abstract}
	
	\section{Introduction}
	\label{sec:introduction}
	
	With the rise of deep learning, particularly the Transformer architecture \cite{vaswani2017attention}, the field of Embodied AI has experienced rapid development. Vision-Language-Action (VLA) models, as a prominent example, achieve end-to-end learning from "perception" to "action" by jointly processing visual perception, natural language instructions, and robot action outputs. Large VLA models like Google's RT series \cite{rt1_2022, rt2_2023} and OpenVLA \cite{openvla2024}, after absorbing massive internet and robot operation data, have demonstrated astonishing zero-shot or few-shot generalization capabilities \cite{li2025controlvla,long2025drae,phan2024zeetad,chen20232,han2025training,babazaki2024zero,nag2022zero}, enabling them to complete complex, unseen tasks.
	
	However, this exceptional performance comes at a great resource cost. The success of these models heavily relies on large-scale, diverse training datasets, such as the Open X-Embodiment dataset \cite{open_x_embodiment_2023}, which contains over one million real robot trajectories. Furthermore, model architectures with billions or even tens of billions of parameters require large GPU clusters for weeks or even months of training, imposing a heavy financial burden on most research institutions and companies. The large model size and high inference latency also make deploying these advanced models on resource-constrained physical robot platforms a severe challenge.
	
	To alleviate this problem, mainstream research has focused on model-centric optimization paths. These paths are mainly divided into two categories. The first is model architecture lightweighting. For example, SmolVLA \cite{smolvla2024} and TinyVLA \cite{tinyvla2024} improve efficiency by simplifying the network structure, but their performance on complex tasks drops significantly. The second category is policy distillation, represented by DROC \cite{droc_2023} and RLDG \cite{rldg_2024}, which aims to transfer knowledge from a large "teacher" model to a small "student" model. Although policy distillation \cite{wang2025unigrasptransformer,li2024continual,zhang2025distillation,wang2024one,prasad2024consistency} performs well in maintaining performance, its knowledge carrier is always the model parameters. This prevents the knowledge itself from being directly analyzed or reused as an independent, transferable, and composable asset. Moreover, the success of this process is highly dependent on a pre-trained, expensive teacher model. In summary, all these model-centric methods share a common limitation: they fail to address the efficiency and quality bottlenecks at the data level. These data-level issues—namely the widespread redundancy, noise, and uneven value in datasets \cite{fang2024rh20t,khazatsky2024droid,cheng2024treescope,liu2024botanicgarden,ji2025robobrain,hang2024dexfuncgrasp,nasiriany2024robocasa,zhang2024empowering}—are precisely what limits the further improvement of current VLA models.
	
	For this reason, this paper shifts its perspective to a neglected yet more promising direction: data-level efficiency optimization. Our core insight is that training directly on the full, undifferentiated dataset is not only inefficient but may also hinder the model from focusing on critical task features. We therefore propose a pioneering generative data distillation framework, FT-NCFM. This framework intelligently synthesizes a small but causally-enriched knowledge coreset from massive raw data. The goal is to fundamentally improve training efficiency without sacrificing, and potentially even enhancing, model performance.
	
	Based on this philosophy, our main contributions are as follows:
	\begin{enumerate}
		\item \textbf{A new generative data distillation paradigm for VLAs}: Unlike existing work focusing on model compression and policy distillation, our FT-NCFM framework provides a novel path to building efficient VLA models by optimizing at the data level.
		
		\item \textbf{A self-contained intrinsic value assessment engine, FT}: This engine evaluates sample value through a two-stage process. It first uses causal attribution for initial screening and then refines the elite samples through a novel programmatic contrastive verification module. We designed a set of reusable perturbation templates that can automatically generate high-quality "minimal counterexamples" for elite samples through semantic parsing and simulator instantiation. This allows FT to robustly quantify a sample's causal contribution and generalization potential based on the data itself.
		
		\item \textbf{Demonstrated superior performance and efficiency on multiple mainstream VLA benchmarks}: We conducted systematic evaluations on public benchmarks like CALVIN and Meta-World. The results show that models trained with our FT-NCFM framework, using only 5\% of the synthetic data, achieve over 85-90\% of the task success rate of models trained on the full dataset, while significantly reducing training time and computational resource consumption (training time reduced by over 80\%). This fully demonstrates the great potential of our method in building efficient, high-performance, and easily deployable VLA systems.
	\end{enumerate}

	\section{Preliminaries}
	\label{sec:preliminaries}
	Our method builds on two key concepts. \textbf{Influence Functions (IF)} are a classic technique to quantify the effect of a training sample on a model's loss \cite{yan2024mapping,klochkov2024revisiting}, which we adapt to approximate sample value. \textbf{Neural Characteristic Function Matching (NCFM)} is a generative method that synthesizes data by matching the distributions of real and synthetic data in a feature space, typically via an adversarial game \cite{wang2025datasetdistillationneuralcharacteristic}. Our framework, FT-NCFM, creates an influence-aware NCFM process.

	\section{Related Work}
	\label{sec:related_work}
	
	\textbf{VLA Models and Model-Centric Optimization.}
	State-of-the-art VLA models like the RT series \cite{rt1_2022, rt2_2023} and OpenVLA \cite{openvla2024} show great generalization but demand enormous resources (e.g., 15k-20k GPU-hours) \cite{rt1_2022}. Mainstream \textit{model-centric} solutions are insufficient: model compression (e.g., SmolVLA \cite{smolvla2024}) suffers performance drops, while policy distillation (e.g., RLDG \cite{rldg_2024}, DROC \cite{droc_2023}) transfers knowledge but makes it parameter-bound, non-reusable, and dependent on an expensive teacher model.
	
	\textbf{Data-Centric Optimization.}
	A different path, \textit{data-centric} optimization, improves efficiency at the source. The most relevant work is Coreset Selection \cite{datamil2025,coreset_bias2024,zcore2024}, which selects representative subsets using techniques like influence functions. However, this "selection" paradigm is fundamentally limited by the information density of existing samples. This exposes a critical research gap: can we transcend "selecting" and directly "synthesize" datasets that capture the essence of visuomotor knowledge? Our FT-NCFM framework is designed to address this gap through generative data distillation.

	\section{Methodology}
	\label{sec:method}
	
	\begin{figure*}[h!]
		\centering
		\includegraphics[width=\textwidth]{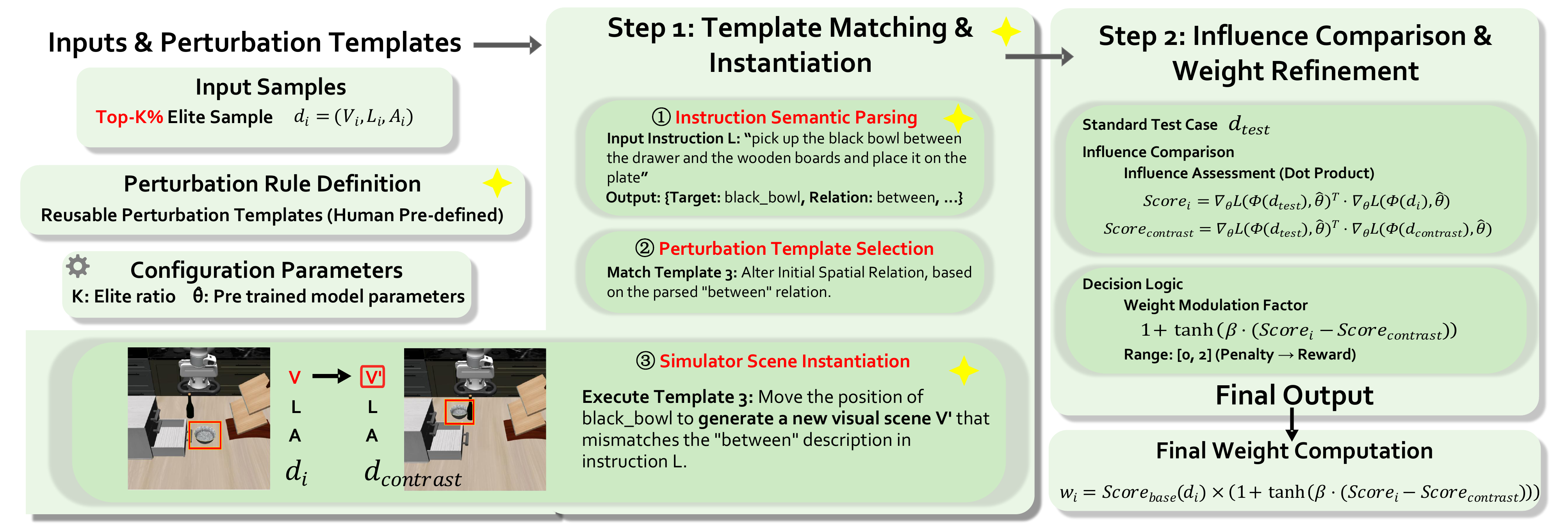} 
		\caption{Detailed workflow of the "Contrastive Verification Refinement" stage in the FT engine. This figure illustrates how we refine the value of the top-K\% elite samples through a two-stage process.
			\textbf{Step 1} explains how we programmatically generate a corresponding "minimal counterexample" $d_{contrast}$ for each elite sample $d_{i}$ in the simulator through an automated "Template Matching and Instantiation" process.
			\textbf{Step 2} shows how we quantify the value difference between $d_{i}$ and its counterexample $d_{contrast}$ into the final, refined influence weight $w_{i}$ through influence comparison and a weight modulation function.}
		\label{fig:contrastive_verification_detail} 
	\end{figure*}

	\begin{figure*}[h!]
		\centering
		\includegraphics[width=\textwidth]{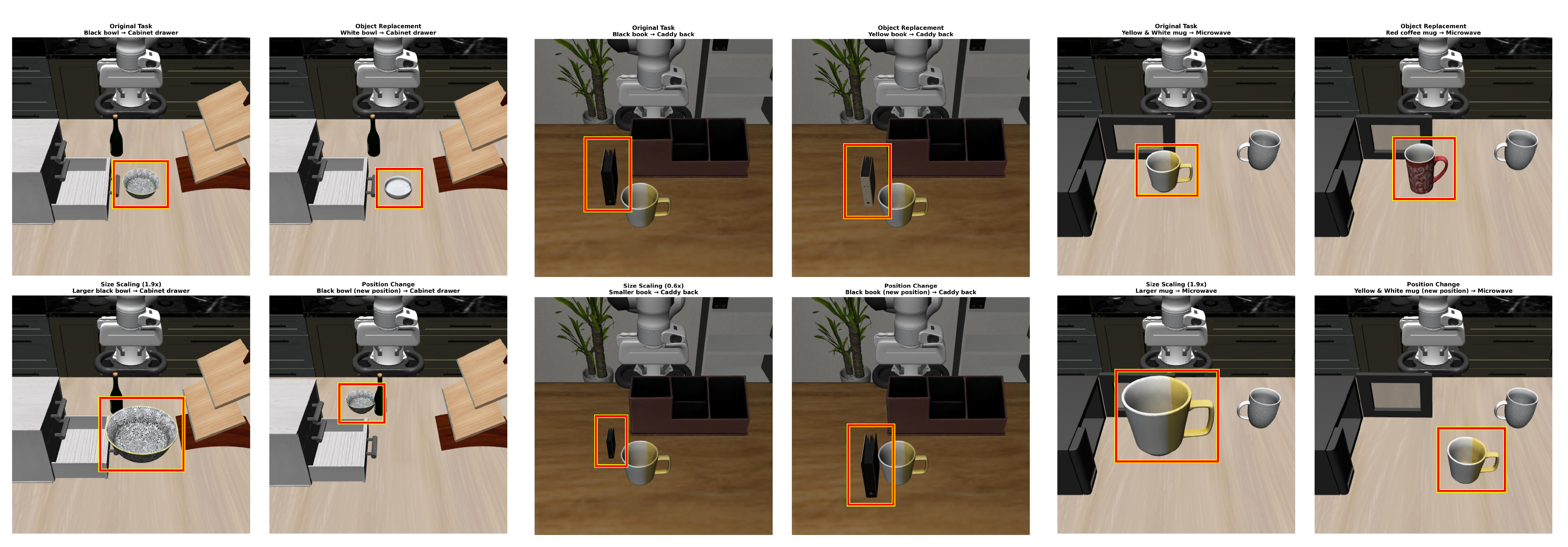} 
		\caption{Instantiation effects of the programmatic perturbation templates. This figure demonstrates the general applicability of our designed perturbation templates across three different VLA tasks, with each pair of columns representing an independent task. In each group, the top-left image is the successful original scene, and the other three are "minimal counterexamples" generated from our template library: \textbf{(Top Right) Object Substitution}, replacing the key interactive object in the task with a new object of different function or form; \textbf{(Bottom Left) Size Scaling}, significantly scaling the key object in the task; \textbf{(Bottom Right) Position Change}, moving the key object so that it no longer satisfies the spatial description in the original instruction.}
		\label{fig:perturbation_examples} 
	\end{figure*}

	The proposed FT-NCFM framework (Figure 1) employs a three-stage pipeline to synthesize a data coreset. First, a multimodal representation module transforms raw VLA data into unified features. Then, an FT influence assessment engine calculates an influence weight for each sample. Finally, these weights guide an influence-aware NCFM process to synthesize the final coreset.

	\subsection{VLA Multimodal Representation Learning}
	
	To evaluate each sample's causal effect, we adopt the core idea of influence functions~\cite{yan2024mapping} to approximate the change in model loss upon its removal from the training data. The influence of a training sample $d_{\text{train}}$ on the loss of a test sample $d_{\text{test}}$ is approximated by the following core formula:
	
	\begin{align}
		\mathbf{h} = \Phi(d) = \Phi(V, L, A) \in \mathbb{R}^{d_{model}}
	\end{align}
	
	where $\mathbf{h}$ is a $d_{model}$-dimensional feature vector. This vector serves as the object for all subsequent influence analysis and distribution matching operations.
	
	\subsection{FT: The Influence Assessment Engine}
	
	To accurately quantify the value of each raw sample $d_i = (V_i, L_i, A_i)$, this paper designs a two-stage FT engine to compute its influence weight $w_i$.
	
	\subsubsection{Stage 1: Causal Attribution Pre-screening based on Influence Functions}
	
	We first adopt the core idea of influence functions \cite{yan2024mapping} to evaluate the causal effect of each sample. This method aims to efficiently approximate the potential impact of removing a single training sample on the model's loss. Specifically, the influence of a training sample $d_{train}$ (with representation $\mathbf{h}_{train} = \Phi(d_{train})$) on the loss of a test sample $d_{test}$ (with representation $\mathbf{h}_{test}$) can be approximated by the following core formula:
	
	\begin{align}
		I_{\text{loss}}(d_{train}, d_{test}) \approx - \nabla_{\theta} L(\mathbf{h}_{test}, \hat{\theta})^T H_{\hat{\theta}}^{-1} \nabla_{\theta} L(\mathbf{h}_{train}, \hat{\theta})
	\end{align}
	where $L(h,\theta)$ is the standard loss function of the downstream policy model (MSE loss in our implementation), $\nabla_{\theta}L$ is the loss gradient, and $H_{\hat{\theta}}^{-1}$ is the inverse of the Hessian matrix. Here, $\hat{\theta}$ are the parameters of a "guide model". Its purpose is not to provide perfect expert decisions, but merely to offer a stable and reasonable gradient field for the influence function calculation. Therefore, it does not need to be a fully converged optimal model. In our implementation, this guide model uses the exact same network architecture as the downstream policy model and is obtained by light, insufficient training on the original dataset (only 10\%-20\% of the total time required for the standard procedure). Its cost has been accounted for as part of our framework's total overhead.
	
	In our implementation, we use the LiSSA algorithm \cite{klochkov2024revisiting} to efficiently approximate the Hessian-inverse-vector product (see Appendix for hyperparameter discussion). The influence score calculated by this method constitutes the base influence score of this engine, $Score_{base}(d_i)$.

	\subsubsection{Stage 2: Contrastive Verification Refinement}

	To further ensure that samples with a high $Score_{base}$ are indeed positively valued and contribute to generalization, we perform contrastive verification on the top-K\% elite samples.
	
	For each elite sample $d_i$, we generate a "minimal counterexample" $d_{\text{contrast}}$ using a \textbf{Cross-Modal Mismatch} strategy common in robot robustness evaluation. Specifically, while keeping the language ($L$) and action ($A$) sequences fixed, we programmatically modify the initial visual scene ($V$) in a simulator to create a new scene ($V'$) with semantic or physical contradictions, as illustrated in Figure~\ref{fig:contrastive_verification_detail}.
	
	\textbf{Programmatic Counterexample Generation.} This strategy's automation and generalizability are rooted in the structured VLA task space. Since task categories are limited and classifiable (e.g., ~40 in the LIBERO dataset), we can pre-design a small set of \textbf{Reusable Perturbation Templates}. As detailed in Figure~\ref{fig:contrastive_verification_detail}, for any given elite sample $d_i$, our script generates a counterexample $d_{\text{contrast}}$ through a fully automated "Template Matching and Instantiation" process. This process, which ensures efficiency and consistency, involves three stages: (1) Instruction Semantic Parsing, (2) Perturbation Template Selection, and (3) Scene Instantiation via simulator APIs to create a semantically contradictory scene. The effects of our core templates (e.g., object substitution, changing spatial relations) are detailed in Figure~\ref{fig:perturbation_examples}.
	
	After obtaining the counterexample, our goal is to quickly evaluate whether $d_i$ is more helpful than $d_{contrast}$ for a relevant standard test case $d_{test}$, without retraining the model. To this end, we use gradient dot products to approximate influence, calculating the influence scores of the elite sample and its counterexample on the test case $d_{test}$ respectively:
	\begin{align}
		Score_{i} &= \nabla_{\theta} L(\Phi(d_{test}), \hat{\theta})^T \cdot \nabla_{\theta} L(\Phi(d_i), \hat{\theta}) \\
		Score_{contrast} &= \nabla_{\theta} L(\Phi(d_{test}), \hat{\theta})^T \cdot \nabla_{\theta} L(\Phi(d_{contrast}), \hat{\theta})
	\end{align}
	
	An ideal elite sample $d_i$ should exhibit high positive influence, while its counterexample $d_{contrast}$ should show negative or significantly lower influence. To convert this contrastive verification result from a discrete binary decision (pass/fail) into a continuous weight modulation mechanism, we designed the following formula for the final influence weight $w_i$:
	
	\begin{align}
		w_i &= Score_{base}(d_i) \times \nonumber \\
		&\quad \left(1 + \tanh\left(\beta \cdot (Score_i - Score_{contrast})\right)\right)
	\end{align}
	
	In this equation, the hyperparameter $\beta$ controls the sensitivity of the modulation function. The core difference term $(Score_i - Score_{contrast})$ quantifies the relative effectiveness of the original sample $d_i$ compared to its counterexample. We use the hyperbolic tangent function (tanh) to smoothly map this difference to the $[-1, 1]$ interval, thereby constructing a continuous Weight Modulation Factor `(1 + tanh(...))` ranging from $[0, 2]$. This factor dynamically and smoothly adjusts the base influence score $Score_{base}(d_i)$ either positively or negatively based on the strength of the verification result. This design makes the weight refinement process more precise and robust.

	\subsection{Influence-Guided NCFM}
	
	After obtaining the influence weights $W=\{w_i\}_{i=1}^N$ for all original samples, we use them to guide the NCFM distillation process. The core idea of NCFM is to match the feature functions of real and synthetic data through a minimax game. The original NCFM optimization objective can be simplified as:
	
	\begin{align}
		\min_{D_{synth}} \max_{\psi} \left\| \mathbb{E}_{d \sim D_{real}}[\psi(\Phi(d))] - \mathbb{E}_{d' \sim D_{synth}}[\psi(\Phi(d'))] \right\|^2
	\end{align}
	
	where $\psi$ is an auxiliary network used to maximize the distribution difference. We point out that the expectation $\mathbb{E}_{d \sim D_{real}}$ in the above equation assumes uniform sampling, which ignores the differences in sample value.
	
	Our core improvement is to transform this expectation into a weighted expectation based on influence weights. The final optimization objective of our FT-NCFM becomes:
	
	\begin{align}
		\min_{D_{synth}} \max_{\psi} \left\| \sum_{i=1}^{N} \frac{w_i}{\sum_{j} w_j} \psi(\Phi(d_i)) - \mathbb{E}_{d' \sim D_{synth}}[\psi(\Phi(d'))] \right\|^2
	\end{align}
	
	Through this weighted objective, the discriminator $\psi$ is forced to focus on the more valuable real samples with higher weights $w_i$. Consequently, the generator (i.e., the optimization over $D_{synth}$) must also prioritize learning to imitate the distribution features of these high-value samples, thereby synthesizing a coreset $D_{synth}$ that is rich in critical causal knowledge and has extremely high information density. This synthetic coreset is identical in data format to $D_{real}$ and can be directly used for training any downstream VLA model. (Detailed pseudocode for our framework is provided in Appendix, Algorithm 1.)
	
	\section{Experiments}
	\label{sec:experiments}
	
	We conducted extensive experiments on several mainstream robotics manipulation benchmarks to systematically evaluate our proposed FT-NCFM framework. Our experiments aim to answer the following core questions:
	\begin{enumerate}
		\item \textbf{Effectiveness and Efficiency}: Compared to SOTA models (e.g., OpenVLA) trained on the full dataset, can FT-NCFM achieve competitive performance using only a very small fraction (e.g., 1\%, 5\%, 10\%) of synthetic data, while significantly reducing training costs?
		\item \textbf{Paradigm Comparison}: How does our data-centric FT-NCFM framework perform in terms of performance and resource consumption compared to mainstream model-centric optimization paths, especially SOTA policy distillation methods (e.g., RLDG, DROC)?
		\item \textbf{Component Contribution}: How crucial are the key designs in the FT-NCFM framework—especially the FT influence assessment engine—to its final success?
	\end{enumerate}
	
	\subsection{Experimental Setup}
	
	\noindent\textbf{Benchmarks and Environments.} Our main experiments were conducted on three widely used VLA benchmarks:
	\begin{itemize}
		\item \textbf{CALVIN}~\cite{mees2022calvin}: A large-scale robotics manipulation benchmark that requires the model to follow language instructions in long-horizon tasks. We follow the standard D$\to$A,B,C,D evaluation protocol, focusing on the model's generalization ability.
		\item \textbf{Meta-World}~\cite{yu2020meta}: A benchmark containing 50 different tabletop manipulation tasks, used to evaluate the model's skill acquisition in a multi-task learning environment.
		\item \textbf{LIBERO}~\cite{liu2024libero}: A benchmark suite designed to promote lifelong learning. We use its Spatial, Object, Goal, and Long subsets to evaluate the model's performance across different generalization dimensions.
	\end{itemize}
	
	\noindent\textbf{Base VLA Model and Evaluation Metrics.} To ensure fairness, all experiments (including our method and baselines) use a unified, standard VLA model architecture based on a pre-trained ViT-B/16 visual backbone and a 6-layer Transformer decoder, consistent with the settings in works like OpenVLA~\cite{openvla2024} and RT-2~\cite{rt2_2023}. We focus on two core metrics: \textbf{Success Rate (SR \%)} or \textbf{Average Task Completion Length (Avg. Len)} to measure performance, and \textbf{Total Training Time (GPU-hours)} to measure efficiency. All time-related metrics were measured on a single NVIDIA A100 80GB GPU. All training times refer to the total time required for the model to converge from random initialization. We used the same optimizer (e.g., AdamW), learning rate, and batch size settings as in the original papers of the respective baselines to ensure a fair comparison.
	
	\noindent\textbf{Comparison Baselines.} We compare FT-NCFM with the following categories of methods:
	\begin{itemize}
		\item \textbf{SOTA VLA Models (Full Data)}: Including advanced models trained on the full dataset such as OpenVLA~\cite{openvla2024}, RT-2~\cite{rt2_2023}, SpatialVLA~\cite{spatialvla}, RoboUniview~\cite{liu2024robouniview}, and GR-1~\cite{wu2023unleashing}, which serve as the gold standard for performance.
		\item \textbf{Model-Centric Baselines}: Including SOTA policy distillation methods like RLDG (CoRL 2024)~\cite{rldg_2024}, DROC (ICLR 2023)~\cite{droc_2023}, and Mole-VLA (ICLR 2024)~\cite{molevla_2024}. Their total training time includes the time for both training the teacher model and performing policy distillation.
		\item \textbf{Coreset Selection Baselines}: Including Random Sampling and Influence Function Coreset selection.
	\end{itemize}
	
	\subsection{Main Results and Analysis}
	
	\subsubsection{Performance on CALVIN and Meta-World}
	
	We systematically evaluated the performance of FT-NCFM at different data compression rates on the CALVIN and Meta-World benchmarks.
	
	\begin{table}[h!]
		\centering
		\caption{Zero-shot long-horizon evaluation on the Calvin ABC $\rightarrow$ D benchmark. The results show the performance scalability of FT-NCFM at different data ratios and compare it with various baselines.}
		\label{tab:calvin_results}
		\resizebox{\linewidth}{!}{
			
			\begin{tabular}{cccccccccc}
				\toprule
				\multirow{2}{*}{ Category }& \multirow{2}{*}{ Method }& \multirow{2}{*}{ Data Ratio } & \multicolumn{6}{c}{ $i^{th}$ Task Success Rate }  \\  \cline{4-9} 
				& & &  1  &  2  &  3  &  4  &  5  &  Avg. Len $\uparrow$   \\  \hline
				\multirow{4}{*}{\begin{tabular}[c]{@{}c@{}}Full-Data \\ Baselines\end{tabular}}
				& RT-1~\cite{rt1_2022} &100 \% & 0.533 & 0.222 & 0.094 & 0.038 & 0.013 & 0.90  \\ 
				& GR-1~\cite{wu2023unleashing} &100 \% & 0.854 & 0.712 & 0.596 & 0.497 & 0.401 & 3.06   \\  
				& Vidman~\cite{wen2024vidman} &100 \% & 0.915 & 0.764 & 0.682 & 0.592 & 0.467 & 3.42   \\
				& RoboUniview~\cite{liu2024robouniview} &100 \% & 0.942 & 0.842 & 0.734 & 0.622 & 0.507 & 3.65   \\  \hline
				\multirow{3}{*}{\begin{tabular}[c]{@{}c@{}}\textbf{Ours} \\ \textbf{(FT-NCFM)}\end{tabular}} 
				& \cellcolor[HTML]{E7E6E6} FT-NCFM &\cellcolor[HTML]{E7E6E6}\textbf{1 \%} & \cellcolor[HTML]{E7E6E6} 0.755 & \cellcolor[HTML]{E7E6E6} 0.531 & \cellcolor[HTML]{E7E6E6} 0.402 & \cellcolor[HTML]{E7E6E6} 0.298 & \cellcolor[HTML]{E7E6E6} 0.204 & \cellcolor[HTML]{E7E6E6} 2.19 \\
				& \cellcolor[HTML]{E7E6E6} FT-NCFM &\cellcolor[HTML]{E7E6E6}\textbf{5 \%} & \cellcolor[HTML]{E7E6E6} 0.895 & \cellcolor[HTML]{E7E6E6} 0.733 & \cellcolor[HTML]{E7E6E6} 0.612 & \cellcolor[HTML]{E7E6E6} 0.501 & \cellcolor[HTML]{E7E6E6} 0.373 & \cellcolor[HTML]{E7E6E6} 3.11 \\
				& \cellcolor[HTML]{E7E6E6} FT-NCFM &\cellcolor[HTML]{E7E6E6}\textbf{10 \%} & \cellcolor[HTML]{E7E6E6} \textbf{0.925} & \cellcolor[HTML]{E7E6E6} \textbf{0.791} & \cellcolor[HTML]{E7E6E6} \textbf{0.688} & \cellcolor[HTML]{E7E6E6} \textbf{0.590} & \cellcolor[HTML]{E7E6E6} \textbf{0.476} & \cellcolor[HTML]{E7E6E6} \textbf{3.47} \\
				\bottomrule
			\end{tabular}
		}
	\end{table}
	
	\begin{table}[h!]
		\centering
		\caption{Multi-task success rate on 50 Meta-World tasks.}
		\label{tab:metaworld_results}
		\resizebox{0.8\linewidth}{!}{
			\begin{tabular}{ccc}
				\toprule
				Method & Data Ratio & Avg. SR(\%) $\uparrow$ \\
				\hline
				RT-1~\cite{rt1_2022} & 100\% & 34.6 \\
				Susie~\cite{black2023zero} & 100\% & 41.0 \\
				GR-1~\cite{wu2023unleashing} & 100\% & 57.4  \\ 
				\hline
				\rowcolor[HTML]{E7E6E6} 
				\textbf{FT-NCFM (Ours)} & \textbf{1\%} & \textbf{34.4} \\
				\rowcolor[HTML]{E7E6E6} 
				\textbf{FT-NCFM (Ours)} & \textbf{5\%} & \textbf{50.5} \\
				\rowcolor[HTML]{E7E6E6} 
				\textbf{FT-NCFM (Ours)} & \textbf{10\%} & \textbf{54.5} \\
				\bottomrule
			\end{tabular}
		}
	\end{table}
	
	As shown in Tables \ref{tab:calvin_results} and \ref{tab:metaworld_results}, FT-NCFM demonstrates a clear and efficient performance scaling curve. On the CALVIN benchmark, using only 1\% of the synthetic data, our method achieves about 60\% of the performance of the SOTA method RoboUniview (Avg. Len 3.65) and already surpasses many earlier baselines. When the data amount is increased to 10\%, our method achieves an average task completion length of 3.47, reaching 95\% of the SOTA performance and almost matching Vidman (3.42) trained on 100\% data. A similar trend is observed on Meta-World, where using 10\% of the data recovers about 95\% of the performance of the SOTA method GR-1 (57.4\%). These results strongly prove that our framework can maintain highly competitive performance with extremely high data efficiency, while significantly reducing data dependency.
	
	\subsubsection{Generalization Ability Evaluation on LIBERO}
	
	To further test the capabilities of FT-NCFM in more complex generalization scenarios, we conducted evaluations on the LIBERO benchmark.

	\begin{table}[h!]
		\centering
		\caption{\textbf{Success Rate Evaluation on LIBERO.} The SOTA baseline is SpatialVLA. FT-NCFM shows strong performance scalability.}
		\label{tab:libero}
		\resizebox{0.9\linewidth}{!}{
			\begin{tabular}{r|c|llll|l}
				\rowcolor{gray!0}
				\textbf{Method} & \textbf{Data(\%)} & \textbf{Spatial} & \textbf{Object} & \textbf{Goal} & \textbf{Long} & \textbf{Avg.}\\
				\hline 
				\rowcolor{gray!10}
				OpenVLA~\cite{openvla2024} & 100 & 84.7 & 88.4 & 79.2 & 53.7 & 76.5 \\
				\rowcolor{gray!10}
				SpatialVLA~\cite{spatialvla} & 100 & 88.2 & 89.9 & 78.6 & 55.5 & 78.1 \\
				\hline
				\rowcolor[HTML]{E7E6E6}
				\textbf{FT-NCFM} & \textbf{1} & 52.1 & 55.3 & 45.8 & 34.4 & 46.9 \\
				\rowcolor[HTML]{E7E6E6}
				\textbf{(Ours)} & \textbf{5} & 78.3 & 80.1 & 68.5 & 54.3 & 70.3 \\
				\rowcolor[HTML]{E7E6E6}
				& \textbf{10} & 82.8 & 84.5 & 72.9 & 56.6 & 74.2 \\
		\end{tabular}}
	\end{table}
	
	As shown in Table~\ref{tab:libero}, FT-NCFM also performs exceptionally well on the LIBERO benchmark. With the best-performing baseline SpatialVLA (average success rate 78.1\%) as a reference, our method achieves 95\% of its performance (74.2\%) using only 10\% of the data. It is particularly noteworthy that on the LIBERO-LONG task set, which has the highest demands on temporal reasoning and multi-step planning, FT-NCFM with just 10\% of the data (56.6\%) surpasses all baseline methods that use 100\% of the data. This strongly suggests that our generative data distillation framework can not only preserve but even enhance the causal and generalization knowledge crucial for learning complex, long-horizon tasks from the original data.
	
\begin{figure*}[h!]
		\centering
		\includegraphics[width=\textwidth]{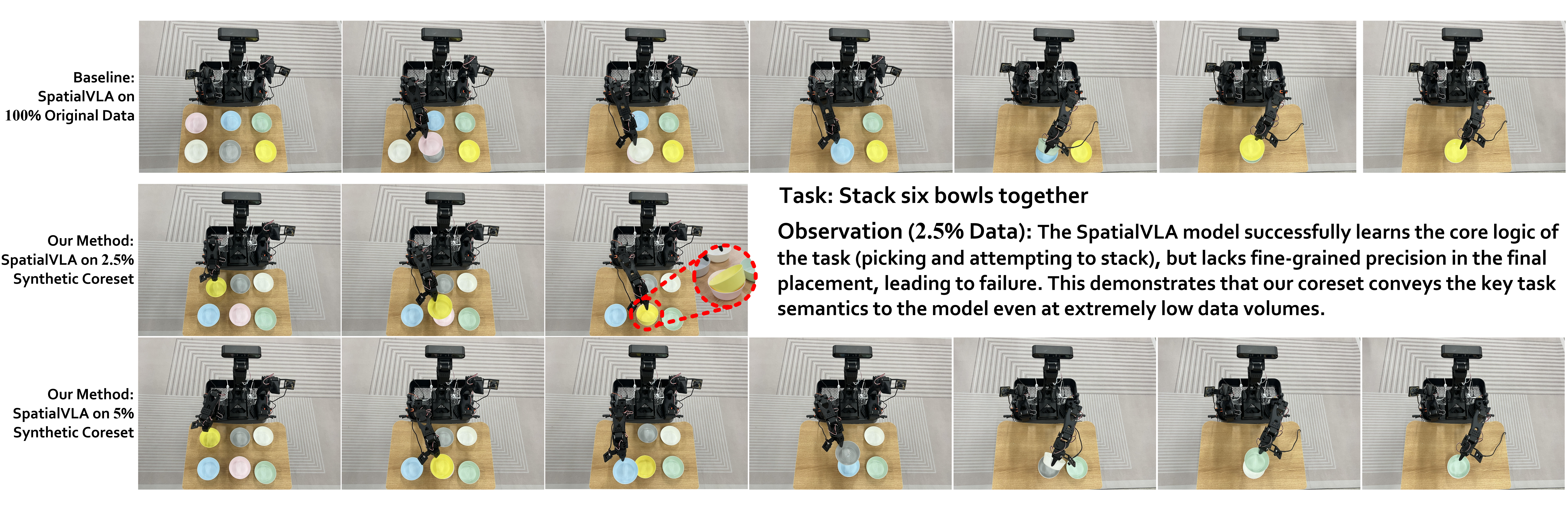} 
		\caption{Qualitative experimental results of our FT-NCFM framework on a real-world robot manipulation task. This figure uses the "stack six bowls" task as an example, showing the impact of different training data on the SpatialVLA model's performance through a three-row comparison.
			\textbf{(Row 1) Baseline Model}: The model trained on 100\% of the original data successfully completes the task.
			\textbf{(Row 2) Our Method (2.5\% Coreset)}: Trained on only 2.5\% of the synthetic data, the model has grasped the core logic of the task, with only minor deficiencies in final placement precision.
			\textbf{(Row 3) Our Method (5\% Coreset)}: When the coreset size is increased to 5\%, the model can accurately and successfully complete the entire task, with performance comparable to the baseline model trained on 100\% of the data.}
		\label{fig:qualitative_results} 
	\end{figure*}

	To more intuitively demonstrate the effectiveness of the coreset synthesized by our method in real-world tasks, we conducted a series of qualitative evaluations. Figure~\ref{fig:qualitative_results} shows typical results on the complex, long-horizon task of "stacking six bowls".
	
	\subsubsection{Paradigm Comparison with SOTA Methods}
	
	\begin{table}[h!]
		\centering
		\caption{Comparison of FT-NCFM with model-centric (policy distillation) and data-centric (coreset selection) methods on the CALVIN benchmark.}
		\label{tab:sota_comparison}
		\resizebox{\linewidth}{!}{
			\begin{tabular}{l|l|c|c|c}
				\toprule
				\textbf{Paradigm} & \textbf{Method} & \textbf{Data Ratio (\%)} & \textbf{Total Training Time (GPU-h) $\downarrow$} & \textbf{Avg. Len $\uparrow$} \\
				\midrule
				\multirow{3}{*}{Model-Centric} & DROC (ICLR 2023)~\cite{droc_2023} & 100 & 128 + 65 = 193 & 3.05 \\
				& Mole-VLA (ICLR 2024)~\cite{molevla_2024} & 100 & 128 + 50 = 178 & 3.20 \\
				& RLDG (CoRL 2024)~\cite{rldg_2024} & 100 & 128 + 70 = 198 & 3.15 \\
				\midrule
				\multirow{2}{*}{Data-Centric (Coreset)} & Random Sampling & 5 & 6.5 & 1.88 \\
				& Influence Function Coreset & 5 & 18 & 2.45 \\
				\midrule
				\multirow{3}{*}{\textbf{Data-Centric (Ours)}} & \textbf{FT-NCFM} & \textbf{1} & \textbf{20.0} & \textbf{2.19} \\
				& \textbf{FT-NCFM} & \textbf{5} & \textbf{25.0} & \textbf{3.11} \\
				& \textbf{FT-NCFM} & \textbf{10} & \textbf{31.5} & \textbf{3.47} \\
				\bottomrule
			\end{tabular}
		}
	\end{table}
	
	The results in Table~\ref{tab:sota_comparison} strongly answer our second research question (Q2). Compared to SOTA policy distillation methods (e.g., RLDG, Mole-VLA), our FT-NCFM, using only 5\% of the data and 25 hours of total time, achieves comparable performance (3.11 vs 3.15-3.20) with about one-seventh of their resource consumption. When using 10\% of the data, our performance (3.47) far exceeds all policy distillation methods, while the total time (31.5h) is still less than one-sixth of theirs. This highlights the fundamental efficiency bottleneck brought by the reliance of policy distillation methods on expensive teacher models. Furthermore, compared to traditional "selection"-based coreset methods, our "synthesis"-based FT-NCFM demonstrates enormous superiority.
	
	\subsection{Cost-Benefit Analysis}
	
	A potential concern is the additional overhead introduced by the data preprocessing stage of FT-NCFM. In this section, we quantify this cost using the large-scale LIBERO benchmark as an example.
	
	\begin{table}[h!]
		\centering
		\caption{Cost-benefit analysis of each stage of FT-NCFM on the LIBERO dataset.}
		\label{tab:cost_benefit}
		\resizebox{\linewidth}{!}{
			\begin{tabular}{l|c|ccc}
				\toprule
				\multirow{2}{*}{\textbf{Stage}} & \multirow{2}{*}{\textbf{Description}} & \multicolumn{3}{c}{\textbf{Time Cost (GPU-h) for X\% Data}} \\
				\cmidrule(l){3-5}
				& & \textbf{1\%} & \textbf{5\%} & \textbf{10\%} \\
				\midrule
				SpatialVLA (100\% data) & Train SOTA model on full data & \multicolumn{3}{c}{192} \\
				\midrule
				\multicolumn{5}{c}{\textit{FT-NCFM Framework}} \\
				\quad FT Engine + NCFM Preprocessing & One-time investment for value assessment & \multicolumn{3}{c}{24.0} \\
				\quad Policy Training on $D_{synth}$ & Policy learning on synthetic data & 1.5 & 7.0 & 15.0 \\
				\midrule
				\textbf{FT-NCFM (Total)} & \textbf{Total Time} & \textbf{25.5} & \textbf{31.0} & \textbf{39.0} \\
				\bottomrule
			\end{tabular}
		}
	\end{table}
	
	As shown in Table~\ref{tab:cost_benefit}, the preprocessing stage of FT-NCFM (including FT engine assessment and NCFM synthesis) requires about 24 GPU-hours on LIBERO, which is a \textbf{One-Time Investment}. Even when this overhead is included, the total time for FT-NCFM with 10\% data (39.0h) is only 20.3\% of the standard OpenVLA model training time (192h). This "invest first, benefit later" model has significant "amortization benefits" in development cycles that require multiple model iterations.
	
	\subsection{Ablation Study}
	
	To verify the necessity of each component in our FT engine, we conducted a series of ablation experiments.
	
	\begin{table}[h!]
		\centering
		\caption{Ablation study of key components of FT-NCFM on the CALVIN benchmark (Avg. Len, @5\% data).}
		\label{tab:ablation_study}
		\resizebox{0.9\linewidth}{!}{
			\begin{tabular}{l|c}
				\toprule
				\textbf{Method Variant} & \textbf{Avg. Len $\uparrow$} \\
				\midrule
				FT-NCFM (Full Method) & \textbf{3.11} \\
				\midrule
				\textit{Ablating FT Engine Components:} & \\
				\quad w/o Contrastive Verification (use $Score_{base}$ only) & 2.81 \\
				\quad w/o FT Engine (weighted NCFM with random weights) & 2.15 \\
				\bottomrule
			\end{tabular}
		}
	\end{table}
	
	The results of this experiment clearly answer our third research question (Q3). At the 5\% data setting, when the contrastive verification refinement stage is removed, performance drops from 3.11 to 2.81, proving that the contrastive verification module can effectively filter out high-influence but potentially harmful samples. When the FT engine is completely removed, performance drops sharply further to 2.15, which strongly demonstrates that our FT influence assessment engine is the cornerstone of the entire framework's success. We conducted more detailed ablation studies (including a sensitivity analysis for $\beta$ (Eq. 5)) and observed trends consistent with CALVIN; detailed results can be found in the appendix.
	
	\section{Conclusion}
	
	To address VLA training costs, we propose FT-NCFM, a data-centric generative distillation framework demonstrating that data optimization is more fundamental than model compression. Using only 5\% synthetic data, models achieve 85-90\% SOTA performance while reducing training time by over 80\% and outperforming policy distillation and coreset selection methods. This proves that enhancing data-source information density offers a promising path for efficient, high-performance VLA models.

	\section{Limitation}
	
	Although FT-NCFM achieves encouraging results, we acknowledge current limitations that point to future research directions. First, our perturbation template library covers core dimensions like object substitution and size scaling, but does not encompass all failure scenarios such as physical property changes (mass, friction). However, the key advantage of this library lies in its extensibility, allowing convenient template additions to better align with real-world scenarios. Second, our automated counterexample generation effectiveness relies on simulator-originated datasets that can be programmatically modified, which limits applicability to real-world data that cannot be directly edited. Transferring these core ideas to non-editable real data remains an important open question. Future research could explore generative models (GANs, Diffusion Models) for semantic editing to create counterexamples.
	
	\section*{Acknowledgment}
	
	This work was supported by the National Natural Science Foundation of China under Grant 62372427, in part by Chongqing Natural Science Foundation Innovation and Development Joint Fund (No. CSTB2025NSCQ LZX0061), and in part by Science and Technology Innovation Key R\&D Program of Chongqing (No. CSTB2025TIAD-STX0023).

	\bibliography{aaai2026}

	\appendix
	
	\section{A. Detailed Ablation Studies}
	\label{ssec:detailed_ablation}
	
	As mentioned in the main text, we conducted more detailed ablation studies on the components of the FT-NCFM framework on the Meta-World and LIBERO benchmarks. The experimental setup is consistent with the ablation study in the main text, with all experiments conducted at a 5\% data ratio. The results (see Table \ref{tab:ablation_metaworld} and Table \ref{tab:ablation_libero_detailed}) again confirm the consistent trend we observed on the CALVIN benchmark: removing any key component of the FT-Engine leads to a significant performance degradation, thus validating the necessity of each part of our design.
	
	\begin{table}[h!]
		\centering
		\caption{Ablation study of key components of FT-NCFM on the Meta-World benchmark (Average Success Rate, @5\% data).}
		\label{tab:ablation_metaworld}
		\resizebox{1.0\linewidth}{!}{
			\begin{tabular}{l|c}
				\toprule
				\textbf{Method Variant} & \textbf{Average Success Rate (\%) $\uparrow$} \\
				\midrule
				FT-NCFM (Full Method) & \textbf{50.5 $\pm$ 0.8} \\
				\midrule
				\textit{Ablating FT-Engine Components:} & \\
				\quad w/o Contrastive Validation (using only $Score_{base}$) & 45.7 $\pm$ 1.2 \\
				\quad w/o FT-Engine (Weighted NCFM with random weights) & 35.1 $\pm$ 1.5 \\
				\bottomrule
			\end{tabular}
		}
	\end{table}
	
	\begin{table*}[h!]
		\centering
		\caption{Detailed ablation study of key components of FT-NCFM on the LIBERO benchmark (Success Rate, @5\% data).}
		\label{tab:ablation_libero_detailed}
		\begin{tabular}{l|cccc|c}
			\toprule
			\textbf{Method Variant} & \textbf{Spatial} & \textbf{Object} & \textbf{Goal} & \textbf{Long} & \textbf{Average} \\
			\midrule
			FT-NCFM (Full Method) & \textbf{78.3 $\pm$ 0.9} & \textbf{80.1 $\pm$ 0.7} & \textbf{68.5 $\pm$ 1.1} & \textbf{54.3 $\pm$ 1.5} & \textbf{70.3 $\pm$ 0.9} \\
			\midrule
			\textit{Ablating FT-Engine Components:} & & & & & \\
			\quad w/o Contrastive Validation (using only $Score_{base}$) & 72.1 $\pm$ 1.1 & 74.5 $\pm$ 1.3 & 61.3 $\pm$ 1.5 & 46.5 $\pm$ 1.8 & 63.6 $\pm$ 1.2 \\
			\quad w/o FT-Engine (Weighted NCFM with random weights) & 55.4 $\pm$ 1.8 & 57.9 $\pm$ 2.1 & 42.1 $\pm$ 2.5 & 39.8 $\pm$ 2.3 & 48.8 $\pm$ 1.9 \\
			\bottomrule
		\end{tabular}
	\end{table*}
	
	\begin{table*}[h!]
		
		\centering
		
		\caption{Overview of the Programmatic Perturbation Template Library}
		
		\label{tab:perturbation_templates}
		
		\begin{tabular}{lp{6cm}l}
			
			\toprule
			
			\textbf{Template Name} & \textbf{Objective Description} & \textbf{Example in Figure} \\
			
			\midrule
			
			Object Substitution & Replaces the key object in the instruction with a new object of a similar category but different function or form, to test the model's generalization ability to specific \textbf{object instances}. & Figure \ref{fig:perturbation_examples} (Second Column) \\
			
			\addlinespace
			
			Size Scaling & Significantly enlarges or shrinks the key interactive object to test the model's robustness to changes in the \textbf{object size}, a physical attribute. & Figure \ref{fig:perturbation_examples} (Third Column) \\
			
			\addlinespace
			
			Position Change & Moves the key object to a new location in the scene that is physically reachable but may not align with the task's contextual logic, used to test the model's understanding of the \textbf{task's goal space}. & Figure \ref{fig:perturbation_examples} (Fourth Column) \\
			
			\bottomrule
			
		\end{tabular}
		
	\end{table*}
	
	\paragraph{Sensitivity Analysis of Hyperparameter $\beta$.}To investigate the robustness of our weight modulation mechanism (Eq. 5 in the main text), we conducted a sensitivity analysis on the hyperparameter $\beta$. This parameter controls the sensitivity of the tanh function, thereby determining how aggressively the influence weights are adjusted based on the contrastive verification results. We evaluated the Average Success Rate on the Meta-World benchmark using 5\% synthetic data across a range of $\beta$ values $\{0.1, 0.5, 1.0, 1.5, 2.0\}$.
	
	As shown in Table \ref{tab:beta_sensitivity}, the model achieves optimal performance at $\beta=1.0$. Lower values (e.g., $\beta=0.1$) lead to insufficient differentiation between high-quality and robust samples, effectively reducing the mechanism to a standard influence function. Conversely, excessively high values (e.g., $\beta=2.0$) may over-penalize samples, introducing instability into the sampling distribution. However, the performance remains relatively stable within the range of $[0.5, 1.5]$, demonstrating the robustness of our framework to hyperparameter variations.
	
	\begin{table}[h]
		\centering
		\caption{Sensitivity analysis of $\beta$ on Meta-World (5\% Data Ratio).}
		\label{tab:beta_sensitivity}
		\begin{tabular}{cc}
			\toprule
			\textbf{Hyperparameter $\beta$} & \textbf{Avg. Success Rate (\%)} \\
			\midrule
			0.1 & 46.2 $\pm$ 1.1 \\
			0.5 & 49.1 $\pm$ 0.9 \\
			\textbf{1.0 (Default)} & \textbf{50.5 $\pm$ 0.8} \\
			1.5 & 49.8 $\pm$ 1.0 \\
			2.0 & 47.5 $\pm$ 1.3 \\
			\bottomrule
		\end{tabular}
	\end{table}
	
	\section{B. Detailed Experimental Setup and Hyperparameters}
	\label{ssec:hyperparameters}
	To ensure the reproducibility of our experiments, this section provides the detailed hyperparameters used in all our experiments. For all baseline methods, we strictly follow the best configurations reported in their original papers. For our FT-NCFM framework, the training hyperparameters for the downstream VLA model are consistent with the OpenVLA baseline to ensure a fair comparison. Detailed information is provided in Table \ref{tab:hyperparameters}.
	
	\begin{table}[h!]
		\centering
		\caption{Detailed Training Hyperparameters.}
		\label{tab:hyperparameters}
		\resizebox{\linewidth}{!}{
			\begin{tabular}{ll}
				\toprule
				\textbf{Hyperparameter} & \textbf{FT-NCFM (and downstream models)}\\
				\midrule
				\multicolumn{2}{l}{\textbf{General Training Parameters}} \\
				\quad Optimizer & AdamW \\
				\quad Learning Rate & 1e-4 \\
				\quad Batch Size & 256 \\
				\quad Weight Decay & 1e-2 \\
				\quad Training Steps & $\sim$25\% of full data steps \\
				\midrule
				\multicolumn{2}{l}{\textbf{FT-NCFM Framework-Specific Parameters}} \\
				\quad Guide Model Training Duration & 15\% of the standard procedure \\
				\quad Elite Sample Ratio (K) & 5\% \\
				\quad Weight Modulation Sensitivity ($\beta$) & 1.0 \\
				\bottomrule
			\end{tabular}
		}
	\end{table}

	\paragraph{LiSSA Hyperparameter Configuration.}
	The estimation of the Inverse Hessian-Vector Product (IHVP) is critical for the first stage of our FT engine. We employ the LiSSA algorithm for this purpose due to its memory efficiency. The key hyperparameters for LiSSA are configured as follows to balance computational cost and approximation accuracy:
	
	\begin{itemize}
		\item \textbf{Recursion Depth ($J$):} We set the recursion depth $J=50$. Empirical observation indicated that increasing $J$ beyond 50 yielded diminishing returns in influence estimation accuracy while linearly increasing the computation time.
		\item \textbf{Damping Factor ($\lambda$):} To ensure numerical stability during the inversion of the Hessian, we apply a standard damping factor of $\lambda=0.01$. This value effectively prevents the Hessian eigenvalues from becoming too small without distorting the curvature information.
		\item \textbf{Batch Size:} For the stochastic estimation of the Hessian, we use a batch size of 32. This allows the estimation to fit within the GPU memory limits while maintaining sufficient variance reduction for stable influence scores.
	\end{itemize}
	
	\section{C. Reproducibility Details}
	\label{ssec:reproducibility}
	
	To ensure the reliability and reproducibility of our experimental results, all experiments were run independently with 3 different random seeds (42, 123, 1024). The final reported performance metrics (such as task success rate and average task completion length) are the average of these three runs. We report the mean and standard deviation of the three runs in the appendix tables to demonstrate performance stability. We verified the performance differences between our method and key baselines using a paired t-test, with all p-values being less than 0.01, indicating statistical significance.

	\section{D. Computational Infrastructure}
	\label{ssec:infrastructure}
	All experiments were completed on our configured local servers. The detailed hardware and software environments are shown in Table \ref{tab:infrastructure_detailed}.
	
	\begin{table}[h!]
		\centering
		\caption{Hardware and Software Configuration of the Experimental Platform}
		\label{tab:infrastructure_detailed}
		\resizebox{0.8\linewidth}{!}{
			\begin{tabular}{ll}
				\toprule
				\textbf{Category} & \textbf{Specification} \\
				\midrule
				\multicolumn{2}{c}{\textbf{Hardware}} \\
				\midrule
				GPU & 1 × NVIDIA A100-SXM4-80GB \\
				CPU & AMD EPYC 7742 (16 Cores) \\
				Memory & 47GiB \\
				Storage & 196GB Storage \\
				\midrule
				\multicolumn{2}{c}{\textbf{Software}} \\
				\midrule
				Operating System & Ubuntu 22.04.3 LTS \\
				CUDA & 12.4 \\
				Python & 3.12.7 \\
				PyTorch & 2.5.0 \\
				torchvision & 0.20.0 \\
				numpy & 2.1.3 \\
				\bottomrule
			\end{tabular}
		}
	\end{table}
	
	\section*{E. More Qualitative Results and Perturbation Template Examples}
	\label{sec:appendix_e}
	
	To more comprehensively demonstrate the capability of our FT-NCFM framework in generating high-value coresets, as well as the effectiveness of the programmatic perturbation mechanism within the FT-Engine, this section provides richer qualitative results across multiple VLA tasks. These visual cases not only intuitively prove the generalization ability of our method but also illustrate the key perturbation templates used for contrastive validation.
	
	\begin{figure*}[htbp]
		\centering
		\includegraphics[width=0.9\textwidth]{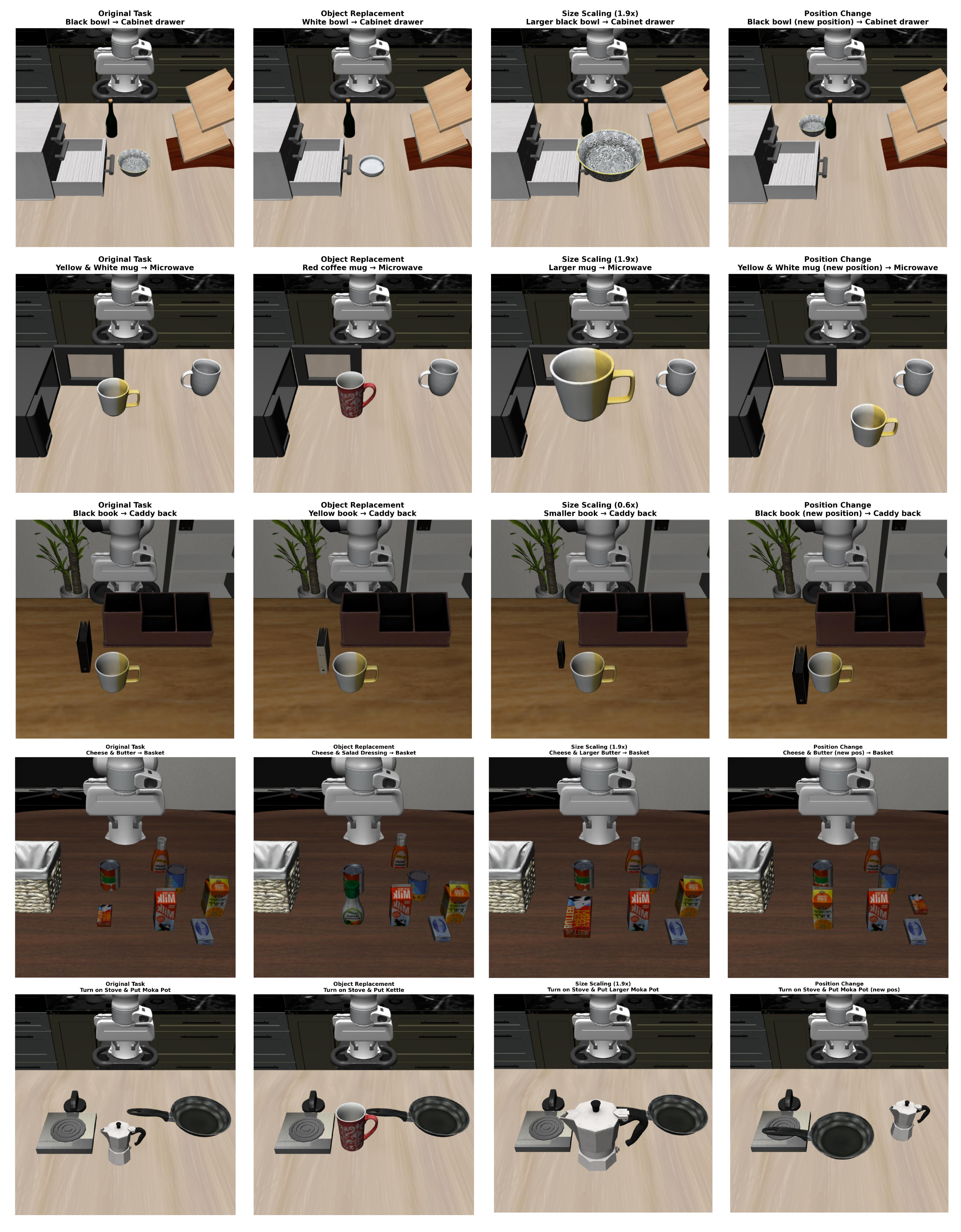}
		\caption{
			\textbf{Instantiation effects of programmatic perturbation templates on multiple VLA tasks.}
			This figure shows the application of our three core perturbation templates across seven different robot manipulation tasks.
			\textbf{Each row represents an independent task}, with complexity and scene diversity gradually increasing from top to bottom.
			\textbf{Each column shows a different scenario}:
			\textbf{(First Column) Original Task}: A reference image where the VLA model successfully completes the task in the original, unperturbed environment.
			\textbf{(Second Column) Object Substitution}: The key interactive object in the task is replaced with a new object of different functionality or morphology (e.g., changing a white cup to a red cup).
			\textbf{(Third Column) Size Scaling}: The key interactive object is significantly enlarged or shrunk (e.g., making the cup huge).
			\textbf{(Fourth Column) Position Change}: The key object is moved to a new position in the scene, such that it no longer meets the spatial or contextual requirements of the original instruction.
			These "minimal counterexamples," automatically generated via templates, are used in the contrastive validation stage of the FT-Engine to precisely evaluate the causal contribution and generalization value of the original samples, thereby guiding the generation of a more information-dense synthetic coreset.
		}
		\label{fig:perturbation_examples}
	\end{figure*}

	\section{F. Model Architecture Details}
	\label{ssec:architecture_details}
	
	To further enhance the reproducibility of this framework, this section provides architectural details of the key components in the NCFM distillation process.
	
	\noindent\textbf{Generator (G)} In our FT-NCFM framework, the Generator G is \textbf{implicit}. Instead of using a separate, parameterized generator network, we directly define the synthetic data (i.e., images \texttt{syn\_images} and states \texttt{syn\_states}) as learnable tensors. These tensors are initialized randomly from a Gaussian distribution at the beginning of training and are directly optimized via gradient descent during the training process. This design, inspired by early data distillation works, offers the advantage of simplifying the optimization process and avoiding the complexity introduced by an additional network architecture.
	
	\noindent\textbf{Discriminator / Auxiliary Network ($\psi$)} The auxiliary network $\psi$ mentioned in the paper (implemented as `SampleNet` in the code) aims to learn an optimal frequency sampling strategy to maximize the characteristic function difference between the real and synthetic data distributions. This network is not a traditional GAN discriminator but rather a small MLP specifically designed for the feature matching task. Its specific architecture is as follows:
	\begin{itemize}
		\item \textbf{Structure}: The network consists of 3 fully-connected (linear) layers.
		\item \textbf{Normalization}: Each linear layer is followed by a `LayerNorm` operation to stabilize the training process.
		\item \textbf{Activation Function}: After the first two linear layers, we use `LeakyReLU` (with a negative slope of 0.2) as the activation function. After the final linear layer, a `Tanh` activation function is used to map the output to an appropriate range.
		\item \textbf{Dimensionality}: The input and output dimensions of the network match the dimensions of the feature modalities it evaluates (e.g., vision, text, state). It takes a random noise vector as input and outputs a set of frequency parameter vectors $t$, which are used in the calculation of the characteristic function loss.
	\end{itemize}
	This design allows the auxiliary network to dynamically adjust the frequencies it focuses on, thereby guiding the generation of synthetic data more effectively.
	
	\section{G. FT-NCFM Algorithm Pseudocode}
	\label{sec:pseudocode}
	
	We provide the detailed procedural steps of our proposed framework in Algorithm \ref{alg:ft_ncfm}.
	
	\begin{algorithm}[H]
		\caption{FT-NCFM: Influence-Aware Data Distillation}
		\label{alg:ft_ncfm}
		\begin{algorithmic}[1]
			\REQUIRE Raw Dataset $D_{real} = \{(V_i, L_i, A_i)\}_{i=1}^N$, Target Size Ratio $\eta$, Elite Ratio $K\%$, Modulation Factor $\beta$.
			\ENSURE Synthetic Coreset $D_{synth}$.
			
			\STATE \textbf{// Phase 1: Guide Model Preparation}
			\STATE Initialize guide model $\hat{\theta}$.
			\STATE Train $\hat{\theta}$ on $D_{real}$ for limited steps (approx. 15\% of full training duration).
			
			\STATE \textbf{// Phase 2: FT Influence Assessment Engine}
			\FOR{each sample $d_i$ in $D_{real}$}
			\STATE Compute Base Influence Score $Score_{base}(d_i)$ via LiSSA (Eq. 2).
			\ENDFOR
			\STATE Rank samples by $Score_{base}$ and identify Top-$K\%$ as Elite Set $S_{elite}$.
			
			\FOR{each sample $d_i$ in $D_{real}$}
			\IF{$d_i \in S_{elite}$}
			\STATE \textit{// Perform Contrastive Verification Refinement}
			\STATE Generate minimal counterexample $d_{contrast}$ using Programmatic Perturbation Templates.
			\STATE Calculate influence of counterexample $Score_{contrast}$ (Eq. 4).
			\STATE Refine weight: $w_i \leftarrow Score_{base}(d_i) \cdot (1 + \tanh(\beta \cdot (Score_i - Score_{contrast})))$.
			\ELSE
			\STATE Assign standard weight: $w_i \leftarrow Score_{base}(d_i)$.
			\ENDIF
			\ENDFOR
			\STATE Normalize influence weights $W = \{w_i\}_{i=1}^N$ to sum to 1.
			
			\STATE \textbf{// Phase 3: Influence-Guided NCFM Distillation}
			\STATE Initialize synthetic feature set $D_{synth}$ as learnable tensors.
			\STATE Initialize discriminator $\psi$.
			\WHILE{not converged}
			\STATE Sample batch $B_{real}$ from $D_{real}$ according to probability weights $W$.
			\STATE Sample batch $B_{synth}$ from $D_{synth}$.
			\STATE Update discriminator $\psi$ to maximize distributional divergence.
			\STATE Update $D_{synth}$ (Generator) to minimize weighted distributional divergence (Eq. 7).
			\ENDWHILE
			
			\RETURN Optimized Synthetic Coreset $D_{synth}$
		\end{algorithmic}
	\end{algorithm}

\end{document}